\documentclass[sigconf]{acmart}

\usepackage{booktabs}
\usepackage{multirow}
\usepackage{makecell}
\usepackage{xcolor}
\usepackage{amsmath}
\usepackage{graphicx}
\usepackage{microtype}
\newcommand{\IEEErows}{590{,}540}
\newcommand{\IEEEfraud}{3.50\%}
\newcommand{\IEEErawLinAUPRC}{0.134}
\newcommand{\IEEErawGBDTAUPRC}{0.616}
\newcommand{\IEEErawGBDTAUC}{0.923}
\newcommand{\IEEEpcaLinAUPRC}{0.112}
\newcommand{\IEEEpcaGBDTAUPRC}{0.455}
\newcommand{\IEEEpcaGBDTAUC}{0.882}
\newcommand{\IEEEgaterLinAUPRC}{0.217}
\newcommand{\IEEEgaterGBDTAUPRC}{0.465}
\newcommand{\IEEEgaterGBDTAUC}{0.868}
\newcommand{\IEEErawLEonepc}{0.162}
\newcommand{\IEEErawLEfivepc}{0.129}
\newcommand{\IEEErawLEtenpc}{0.131}
\newcommand{\IEEErawLEtwentyfivepc}{0.136}
\newcommand{\IEEEpcaLEonepc}{0.105}
\newcommand{\IEEEpcaLEfivepc}{0.111}
\newcommand{\IEEEpcaLEtenpc}{0.110}
\newcommand{\IEEEpcaLEtwentyfivepc}{0.114}
\newcommand{\IEEEgaterLEonepc}{0.206}
\newcommand{\IEEEgaterLEfivepc}{0.219}
\newcommand{\IEEEgaterLEtenpc}{0.215}
\newcommand{\IEEEgaterLEtwentyfivepc}{0.219}
\newcommand{\IEEErawLinStd}{0.002}
\newcommand{\IEEErawGBDTStd}{0.006}
\newcommand{\IEEEpcaLinStd}{0.001}
\newcommand{\IEEEpcaGBDTStd}{0.005}
\newcommand{\IEEEgaterLinStd}{0.034}
\newcommand{\IEEEgaterGBDTStd}{0.013}

\newcommand{\TFrows}{2{,}000{,}000}
\newcommand{\TFfraud}{0.13\%}
\newcommand{\TFrawLinAUPRC}{0.013}
\newcommand{\TFrawGBDTAUPRC}{0.022}
\newcommand{\TFrawGBDTAUC}{0.835}
\newcommand{\TFpcaLinAUPRC}{0.012}
\newcommand{\TFpcaGBDTAUPRC}{0.017}
\newcommand{\TFpcaGBDTAUC}{0.818}
\newcommand{\TFfreqLinAUPRC}{0.023}
\newcommand{\TFfreqGBDTAUPRC}{0.411}
\newcommand{\TFfreqGBDTAUC}{0.871}
\newcommand{\TFgaterLinAUPRC}{0.008}
\newcommand{\TFgaterGBDTAUPRC}{0.508}
\newcommand{\TFgaterGBDTAUC}{0.876}
\newcommand{\TFrawLinStd}{0.001}
\newcommand{\TFrawGBDTStd}{0.002}
\newcommand{\TFpcaLinStd}{0.000}
\newcommand{\TFpcaGBDTStd}{0.001}
\newcommand{\TFfreqLinStd}{0.001}
\newcommand{\TFfreqGBDTStd}{0.056}
\newcommand{\TFgaterLinStd}{0.009}
\newcommand{\TFgaterGBDTStd}{0.063}
\newcommand{\TFgaterGBDTlift}{factor of 23} % 0.508 / 0.022

\newcommand{\IEEEablFULLlin}{0.199}
\newcommand{\IEEEablFULLgbdt}{0.492}
\newcommand{\IEEEablNOCONTlin}{0.151}
\newcommand{\IEEEablNOCONTgbdt}{0.467}
\newcommand{\IEEEablNOEXClin}{0.186}
\newcommand{\IEEEablNOEXCgbdt}{0.425}
\newcommand{\IEEEablFEATlin}{0.241}
\newcommand{\IEEEablFEATgbdt}{0.469}
\newcommand{\IEEEablGRUlin}{0.212}
\newcommand{\IEEEablGRUgbdt}{0.494}
\newcommand{\IEEEablMLPlin}{0.203}
\newcommand{\IEEEablMLPgbdt}{0.499}

\newcommand{\IEEEablNOKLlin}{0.208}
\newcommand{\IEEEablNOKLgbdt}{0.481}

\newcommand{\IEEEablNOKLLEonepc}{0.215}
\newcommand{\IEEEablFULLLEonepc}{0.219}

\newcommand{\TFablFULLlin}{0.014}
\newcommand{\TFablFULLgbdt}{0.401}
\newcommand{\TFablNOCONTlin}{0.005}
\newcommand{\TFablNOCONTgbdt}{0.365}
\newcommand{\TFablNOEXClin}{0.031}
\newcommand{\TFablNOEXCgbdt}{0.513}
\newcommand{\TFablFEATlin}{0.009}
\newcommand{\TFablFEATgbdt}{0.533}
\newcommand{\TFablGRUlin}{0.015}
\newcommand{\TFablGRUgbdt}{0.510}
\newcommand{\TFablMLPlin}{0.008}
\newcommand{\TFablMLPgbdt}{0.487}

\newcommand{\TFablNOKLlin}{0.219}
\newcommand{\TFablNOKLgbdt}{0.431}

\newcommand{\IEEEcatboostGBDTAUPRC}{0.508}
\newcommand{\IEEEcatboostGBDTAUC}{0.907}
\newcommand{\IEEEcatboostLEonepc}{0.348}
\newcommand{\TFcatboostGBDTAUPRC}{0.711}
\newcommand{\TFcatboostGBDTAUC}{0.982}

\newcommand{\IEEEexcCellLin}{0.227}
\newcommand{\IEEEexcCellLinStd}{0.039}

\newcommand{\IEEEexcFeatLin}{0.217}
\newcommand{\IEEEexcFeatLinStd}{0.034}

\newcommand{\IEEEexcScarfLin}{0.197}
\newcommand{\IEEEexcScarfLinStd}{0.022}

\newcommand{\RCPCALin}{0.112}
\newcommand{\RCPCAGbdt}{0.455}
\newcommand{\RCPCASil}{0.152}
\newcommand{\RCPCALEonepc}{0.105}
\newcommand{\RCPCATrain}{$<1$}
\newcommand{\RCUMAPLin}{0.050}
\newcommand{\RCUMAPGbdt}{0.357}
\newcommand{\RCUMAPSil}{0.202}
\newcommand{\RCUMAPLEonepc}{0.046}
\newcommand{\RCUMAPLEfivepc}{0.049}
\newcommand{\RCUMAPLEtenpc}{0.049}
\newcommand{\RCUMAPLEtwentyfivepc}{0.050}
\newcommand{\RCUMAPTrain}{1{,}318}
\newcommand{\RCAELin}{0.179}
\newcommand{\RCAEGbdt}{0.468}
\newcommand{\RCAESil}{0.215}
\newcommand{\RCAELEonepc}{0.147}
\newcommand{\RCAETrain}{319}
\newcommand{\RCSAELin}{0.139}
\newcommand{\RCSAEGbdt}{0.446}
\newcommand{\RCSAESil}{0.225}
\newcommand{\RCSAELEonepc}{0.135}
\newcommand{\RCSAETrain}{303}
\newcommand{\RCVVAELin}{0.083}
\newcommand{\RCVVAEGbdt}{0.443}
\newcommand{\RCVVAESil}{0.717}
\newcommand{\RCVVAELEonepc}{0.083}
\newcommand{\RCVVAETrain}{286}
\newcommand{\RCSCARFLin}{0.214}
\newcommand{\RCSCARFGbdt}{0.491}
\newcommand{\RCSCARFSil}{0.086}
\newcommand{\RCSCARFLEonepc}{0.217}
\newcommand{\RCSCARFTrain}{788}
\newcommand{\RCGATERLin}{0.212}
\newcommand{\RCGATERGbdt}{0.450}
\newcommand{\RCGATERSil}{0.081}
\newcommand{\RCGATERLEonepc}{0.187}
\newcommand{\RCGATERTrain}{506}
\newcommand{\RCgaterTimeFracScarf}{64\%} % GATER train time as % of SCARF's
\newcommand{\RCscarfTimeExtra}{56\%} % SCARF trains this % longer than GATER
\newcommand{\RCgaterPcaLinRatio}{1.9$\times$} % GATER lin-probe vs PCA

\newcommand{\ClusterARI}{0.9991}
\newcommand{\ClusterARIstd}{0.0005}

\newcommand{\EPkbest}{3}
\newcommand{\EPsil}{0.085}
\newcommand{\EPConefraud}{8.4\%}

\newcommand{\EPCthreefraud}{2.1\%}
\newcommand{\EPConesize}{103K}

\newcommand{\EPfraudratio}{4.0$\times$}

\newcommand{\FTTonepc}{0.253}

\newcommand{\FTTfull}{0.696}

\newcommand{\best}[1]{\textbf{#1}}

\copyrightyear{2026}
\acmYear{2026}
\setcopyright{cc}
\setcctype{by}
\acmConference[ICAIF '26]{7th ACM International Conference on AI in Finance}{November
14--17, 2026}{Milan, AA, Italy}
\acmBooktitle{7th ACM International Conference on AI in Finance (ICAIF '26), November
14--17, 2026, Milan, AA, Italy}
\acmDOI{10.1145/3858063.3858708}
\acmISBN{979-8-4007-3087-0/2026/11}
\begin{document}

\title{CARTA: Self-Supervised Reusable Transaction Embeddings via
       Heterogeneous Tokenization and Cross-Attention}

\author{Shreyas K.\ Chandrahas}
\authornote{Corresponding author.}
\affiliation{%
  \institution{Visa Inc.}
  \city{Foster City}
  \state{CA}
  \country{USA}}
\email{shrchand@visa.com}

\author{Jose Munoz}
\affiliation{%
  \institution{Massachusetts Institute of Technology}
  \city{Cambridge}
  \state{MA}
  \country{USA}}
\email{josemm@mit.edu}

\author{Satish Babu Bellapu}
\affiliation{%
  \institution{Visa Inc.}
  \city{Foster City}
  \state{CA}
  \country{USA}}
\email{sbellapu@visa.com}

\begin{abstract}
Acquiring fraud labels for payment transactions is expensive and inconsistent
across tasks, and high-cardinality identifiers such as merchant names make
one-hot feature pipelines infeasible at scale. We present CARTA, a
self-supervised encoder that learns a compact 16-dim embedding from raw
transactions without labels; the frozen embedding is then reused for
label-efficient fraud probing and exploratory unsupervised risk
stratification without retraining the encoder.
CARTA uses type-specific projections with learned embedding tables for
high-cardinality identifiers, trained with token excision and InfoNCE
contrastive loss. On IEEE-CIS, its linear probe at 1\% of labels exceeds a
raw-feature probe at 100\%; on IBM TabFormer (2M rows), its frozen embedding
reaches GBDT AUPRC $\TFgaterGBDTAUPRC$, above frequency encoding
($\TFfreqGBDTAUPRC$) and the merchant-dropping baseline ($\TFrawGBDTAUPRC$),
though fully-supervised CatBoost is stronger. CARTA is a label-free, reusable
representation that complements task-specific supervised models.
\end{abstract}



\maketitle

% ─────────────────────────────────────────────────────────────────────────────
\section{Introduction}
\label{sec:intro}

Banks and payment networks authorize billions of card transactions, and
flagging the fraudulent ones is a core analytics task. The signal that
separates fraud from legitimate spending seldom lies in any single field.
For example, a \$500 purchase is unremarkable on its own, but the same amount
at an atypical merchant, at an odd hour, from a rarely-seen device can be
highly suspicious.
Useful models must therefore capture interactions among heterogeneous
fields, often several at once.

Each authorization record is correspondingly high-dimensional and heterogeneous,
spanning continuous monetary amounts (transaction value, settlement amount),
low-cardinality flags (card type, payment method), timestamps
(hour of day, day of week), and high-cardinality identifiers such as merchant
name, city, and postal code. Encoding the categorical fields by one-hot encoding
(OHE) inflates a single record to several hundred dimensions, and for the
high-cardinality fields, with thousands of distinct values each, OHE becomes
infeasible outright: one-hot encoding the merchant columns of a 2M-row benchmark
yields a $\sim$6{,}600-dimensional matrix that exceeds available memory.

Acquiring labeled fraud examples is expensive, delayed, and inconsistent
across analytical tasks. Fraud labels typically arrive days after the
transaction, rule-based labeling introduces noise, and the same transaction
record that drives fraud scoring must also feed behavioral drift monitors
and forecasting systems, each requiring its own bespoke feature-engineering
pipeline. This per-task engineering must be reworked when new feature columns
are added or removed, and creates inconsistent representations across the
analytical stack.

We propose CARTA (Cross-Attention Reusable Transaction Autoencoder), a
self-supervised system that learns once from raw transaction data and produces
a single 16-dimensional embedding reusable across downstream tasks without
retraining. It routes each feature type through an appropriate projection:
continuous features through a learned dense layer, low-cardinality categoricals
through a one-hot projection, and high-cardinality identifiers through learned
embedding tables. The encoder is a Perceiver-style
architecture~\cite{jaegle2021perceiver} whose fixed set of latent query vectors
cross-attend to the feature tokens, decoupling computational cost from input
dimensionality; GRU-gated and MLP variants are evaluated as alternatives
(Section~\ref{sec:encoder}).

The training methodology combines three elements:
(i)~\emph{Token excision}, a self-supervised corruption augmentation that
masks a random subset of feature tokens per view, so the encoder must rely on
the surviving context (on IEEE-CIS this performs on par with SCARF-style
corruption); (ii)~an InfoNCE
contrastive loss~\cite{oord2018representation} on differently-excised views
of the same transaction, making the latent space discriminative; and
(iii)~cyclical KL annealing to prevent the posterior collapse that afflicts
vanilla VAEs~\cite{kingma2013auto}.

\textit{Contributions.}
Each component (Perceiver encoders, InfoNCE, VAE training, learned embeddings
for high-cardinality categoricals) has precedent in isolation; their co-design
for payment-transaction data is what yields properties unavailable from any
single prior method. Concretely:
\begin{enumerate}
  \item \textit{Heterogeneous tokenization for finance}: a type-specific
        projection layer (dense for continuous, OHE for low-cardinality,
        learned embedding tables~\cite{guo2016entity} for high-cardinality identifiers) that avoids
        one-hot explosion without discarding merchant-identity signal.
  \item \textit{A single reusable, label-free embedding}: one frozen 16-dim
        per-record representation evaluated across linear and GBDT probes,
        label-efficiency, and unsupervised clustering, benchmarked against
        PCA, UMAP, autoencoders and SCARF (Table~\ref{tab:repcomp}), with
        full-label linear-probe quality similar to SCARF's at a shorter
        single-run training time.
  \item \textit{Label-efficiency evidence}: on a linear probe, the standard
        self-supervised measure of representation quality, CARTA at 1\% of
        labels exceeds raw features at 100\% of labels; gradient-boosted decision
        trees (GBDT) on raw features (Table~\ref{tab:main}) give a strong
        fully-supervised baseline for reference.
  \item An \textit{ablation study} that evaluates the InfoNCE objective
        (removed jointly with its paired-view augmentation), the
        token-excision mode, and cyclical KL annealing, and compares three
        encoder architectures (Perceiver, GRU-gated, MLP), including cases where
        simpler alternatives are competitive.
\end{enumerate}

% ─────────────────────────────────────────────────────────────────────────────
\section{Related Work}
\label{sec:related}

\textit{Sequential transaction transformers.}
TabBERT and TabFormer~\cite{padhi2021tabular} apply masked-language-model
pretraining over sequences of transactions, using a hierarchical transformer
to capture both intra-record field interactions and inter-record temporal
patterns. FATA-Trans~\cite{zhang2023fata} extends this with field- and
time-aware attention. These methods produce per-position sequence
representations and are evaluated on a single downstream head; they do not
produce a compact per-record embedding reusable across analytical tasks, and
their O($N^2$) self-attention scales quadratically with sequence length.
CARTA produces a fixed 16-dim per-record embedding in O($Q{\cdot}N$)
where $Q=8$ latent queries are independent of input dimensionality.

\textit{Self-supervised tabular learning.}
VIME~\cite{yoon2020vime} predicts mask vectors from corrupted inputs.
SubTab~\cite{ucar2021subtab} splits features into overlapping subsets and
combines reconstruction with contrastive loss. SCARF~\cite{bahri2022scarf}
randomly corrupts individual feature values and applies SimCLR-style
contrastive learning. CARTA uses token excision in the same spirit as these
corruption schemes; in a controlled 5-training-seed comparison on IEEE-CIS
(Section~\ref{sec:ablation})
cell-, feature-, and SCARF-style corruption perform comparably, so we treat the
corruption scheme as a standard SSL component rather than a distinguishing one.
TabTransformer~\cite{huang2020tabtransformer}, the closest prior, pretrains
contextual categorical embeddings via self-attention but yields no compact
reusable embedding. What CARTA adds over all of these is the combination of
learned high-cardinality embedding tables with a compact, reusable per-record
latent for payment-transaction data.

\textit{Supervised tabular transformers.}
FT-Transformer~\cite{gorishniy2021revisiting} tokenizes all features
(continuous and categorical) and applies standard transformer self-attention,
achieving strong results on supervised benchmarks. SAINT~\cite{somepalli2021saint}
extends this with inter-sample attention, and TabNet~\cite{arik2021tabnet} uses sequential
attention for feature selection. These are supervised architectures that require full
label access and retraining per task; they do not produce reusable,
task-agnostic self-supervised embeddings. For fraud detection specifically,
\citet{fu2024challenging} show that tabular transformers can approach GBDT
performance with sufficient data.

\textit{Tabular foundation models.}
Cross-table pretrained models target transfer across datasets:
TabPFN~\cite{hollmann2025tabpfn} (in-context inference), XTab~\cite{zhu2023xtab}
(a shared cross-table transformer), and CARTE~\cite{kim2024carte} (graph
attention over cell strings for high-cardinality entities). These emphasize
cross-table generalization; CARTA instead learns one frozen, reusable per-record
embedding for a single transaction stream. CARTE is closest for high-cardinality
strings but targets cross-table transfer, not a reusable per-stream
representation.

% ─────────────────────────────────────────────────────────────────────────────
\section{Method}
\label{sec:method}

\begin{figure*}[t]
\centering
\includegraphics[width=\textwidth]{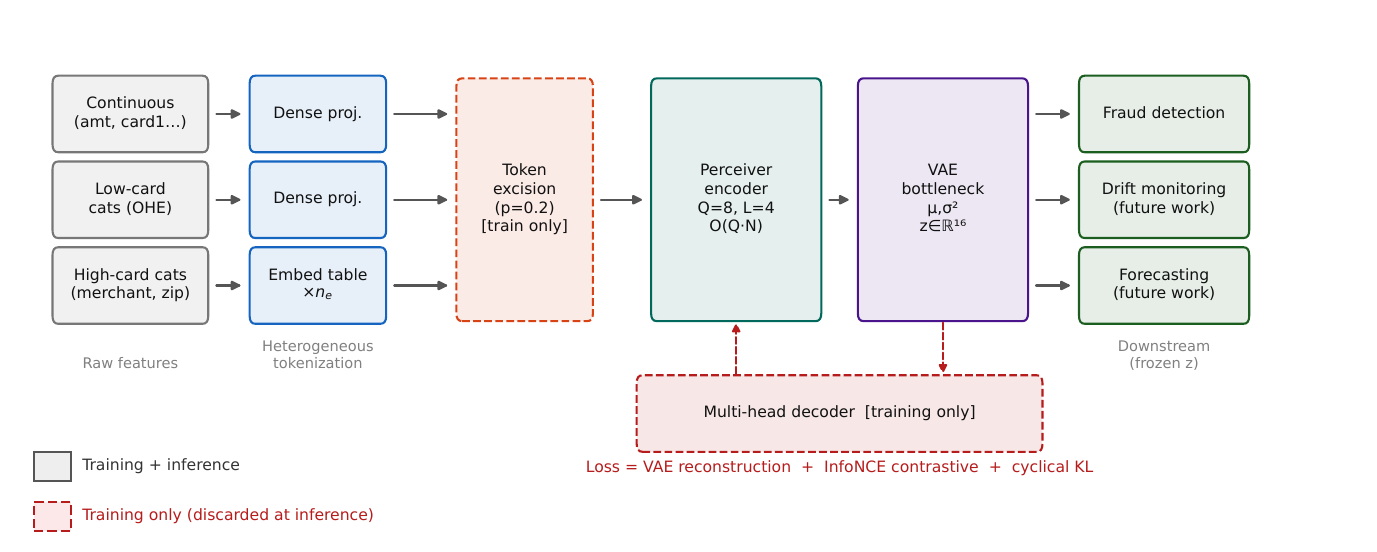}
\caption{CARTA system overview. Raw transaction features are projected into a
uniform token space via type-specific transformations (continuous features to a
dense projection, low-cardinality categoricals to a one-hot projection, and
high-cardinality identifiers to learned embedding tables). Token excision randomly masks a
subset of feature tokens per view as a self-supervised corruption. The Perceiver encoder compresses
tokens into a 16-dim Variational Autoencoder (VAE)~\cite{kingma2013auto} latent.
Training uses a composite loss ($\mathcal{L}_\text{VAE}+\mathcal{L}_\text{InfoNCE}$ with cyclical KL). At
inference the frozen embedding is reused across downstream tasks.}
\label{fig:arch}
\end{figure*}

Figure~\ref{fig:arch} overviews the pipeline: type-specific token projections,
token excision into two views, a Perceiver encoder with a 16-dim VAE latent, and
a composite VAE\,$+$\,InfoNCE objective, after which the frozen latent is reused
across tasks.

\subsection{Heterogeneous Feature Tokenization}
\label{sec:tokenization}

Let a transaction record contain $n_c$ continuous features, $n_o$
low-cardinality categorical features (tractable for one-hot encoding), and
$n_e$ high-cardinality categorical features (vocabularies too large for practical one-hot encoding, i.e., hundreds to thousands of distinct values).
CARTA converts each group to a uniform $d$-dimensional token space:
\begin{align}
\mathbf{t}^{\text{cont}} &= \mathbf{W}_c \mathbf{x}_c \in \mathbb{R}^d,
    \quad \text{(one token)} \\
\mathbf{t}^{\text{ohe}} &= \mathbf{W}_o \text{OHE}(\mathbf{x}_o) \in \mathbb{R}^d,
    \quad \text{(one token)} \\
\mathbf{t}^{(i)}_{\text{emb}} &= \mathbf{E}_i[x_e^{(i)}] \in \mathbb{R}^d,
    \quad i=1,\ldots,n_e. \quad \text{(one token each)}
\end{align}
The resulting token sequence
$\mathbf{X} = [\mathbf{t}^{\text{cont}}, \mathbf{t}^{\text{ohe}},
\mathbf{t}^{(1)}_\text{emb}, \ldots, \mathbf{t}^{(n_e)}_\text{emb}]
\in \mathbb{R}^{N \times d}$ with $N = 2 + n_e$
is augmented with learnable positional embeddings.
Prior to tokenization, continuous features with heavy-tailed
power-law distributions (detected via log-likelihood ratio
test~\cite{clauset2009power}) are transformed with $\log(x+1)$ followed
by Box-Cox normalization.

\subsection{Perceiver Encoder and Variational Bottleneck}
\label{sec:encoder}

The core encoder adapts the Perceiver architecture~\cite{jaegle2021perceiver}
to tabular data. A set of $Q=8$ learnable latent query vectors
$\mathbf{Q} \in \mathbb{R}^{Q \times d}$ cross-attend to the feature tokens
at each of $L=4$ encoder layers:
\begin{equation}
\mathbf{Q} \leftarrow \text{LayerNorm}\!\left(\mathbf{Q} +
  \text{Dropout}\!\left(\text{MHA}(\mathbf{Q}, \mathbf{X}, \mathbf{X})\right)\right),
\end{equation}
where $\text{MHA}$ is multi-head dot-product attention with 4 heads.
Self-attention among latent queries, followed by a two-layer feed-forward
network with GELU activations, allows the queries to specialize.
After the final layer, the $Q$ latent vectors are mean-pooled to a single
context vector $\mathbf{c} \in \mathbb{R}^d$, which is projected to the
variational posterior parameters $(\boldsymbol{\mu}, \log\boldsymbol{\sigma}^2)
\in \mathbb{R}^{z}$ with $z=16$.

The cross-attention complexity is $O(Q \cdot N)$ rather than $O(N^2)$;
since $Q=8$ is fixed, cost is linear in the number of input features.

\textit{GRU-gated variant.}
An alternative encoder replaces the $Q$ latent queries with a single context
vector $\mathbf{c} \in \mathbb{R}^d$ iteratively refined across $L$ layers:
at each layer the context cross-attends to the feature tokens, and the
attention output is integrated via a GRU cell~\cite{cho2014learning} with
learned update and reset gates. This variant is more memory-efficient
and is compared empirically in Section~\ref{sec:ablation}.

\subsection{Token Excision and Hybrid Training Objective}
\label{sec:training}

\textit{Token excision.}
During training, two corrupted views are created per transaction ($p{=}0.2$).
\emph{Cell mode} independently zeroes individual continuous values and
individual one-hot elements (element-wise Bernoulli masking before
tokenization), leaving embedding indices unmasked (zeroing an integer index
would map to a different category).
\emph{Feature mode} (default) masks whole feature groups in the input, also
before tokenization: the continuous group is zeroed as a unit, each one-hot
field's block is zeroed as a unit, and embedding indices are replaced by index~0
($\langle\text{unknown}\rangle$), so corruption reaches
high-cardinality embedding features. Section~\ref{sec:ablation} confirms all
three modes are comparable on IEEE-CIS; feature mode is default because it is
the only variant that masks embedding features.

\textit{Composite loss.}
The two excised views $\mathbf{v}_1, \mathbf{v}_2$ are independently encoded
to latent means $\mathbf{z}_1, \mathbf{z}_2$. The total loss is:
\begin{equation}
\mathcal{L} = \tfrac{1}{2}[\mathcal{L}_\text{VAE}(\mathbf{v}_1)
  + \mathcal{L}_\text{VAE}(\mathbf{v}_2)]
  + \lambda_c \, \mathcal{L}_\text{InfoNCE}(\mathbf{z}_1, \mathbf{z}_2),
\end{equation}
where $\mathcal{L}_\text{VAE} = \mathcal{L}_\text{recon} + \beta\,
D_\text{KL}(q(\mathbf{z}|\mathbf{x})\,\|\,p(\mathbf{z}))$ and the
reconstruction loss is type-appropriate: L2 for continuous features; a
\emph{selected-index} softmax cross-entropy over the concatenated
low-cardinality one-hot block, whose single integer target is the argmax of
the concatenation --- so only that selected index is directly supervised per
record, not every low-cardinality field (a limitation of the implementation
used for all reported results, under which the remaining low-cardinality
fields are reconstructed only indirectly); and a separate per-vocabulary
softmax cross-entropy for each high-cardinality embedding field.
The InfoNCE loss treats the two views of the same transaction as a positive
pair and all other transactions in the batch as negatives (temperature
$\tau = 0.07$, weight $\lambda_c = 0.1$).

\textit{Cyclical KL annealing.}
To prevent posterior collapse, $\beta$ follows a cyclical schedule
over $C=10$ cycles, linearly increasing from $0$ to $\beta_\text{max}=1$
during the first half of each cycle. This allows the encoder to periodically
escape collapse by temporarily reducing the KL penalty~\cite{fu2019cyclical}.

\subsection{Decoder and Downstream Use}
\label{sec:decoder}

An MLP decoder maps $\mathbf{z}$ back to the original feature space via
type-specific output heads: L2 loss for continuous features; a shared logits
head over the concatenated one-hot block trained with the selected-index
cross-entropy of Section~\ref{sec:training}; and $n_e$ separate per-vocabulary
heads for the high-cardinality features, each with its own cross-entropy. After
training, only the encoder is deployed; the decoder is discarded. Downstream
tasks consume the frozen 16-dim embedding vector $\boldsymbol{\mu}$
(the posterior mean of the VAE), requiring no retraining.

% ─────────────────────────────────────────────────────────────────────────────
\section{Experimental Setup}
\label{sec:setup}

\subsection{Datasets}
\textit{IEEE-CIS Fraud}~\cite{ieeecis2019} (\IEEErows{} rows, \IEEEfraud{}
fraud): a widely used public card-transaction fraud benchmark. From its 400+
largely anonymized columns we retain 15 numeric and 3 low-cardinality
categorical fields with low missingness (the full list is in the accompanying
code); this retained subset has no high-cardinality identifiers, so the GBDT number here
is an upper bound over this feature subset, not over all 400+ raw columns. Labels
are known throughout but withheld from the encoder during pretraining.

\textit{TabFormer} (\TFrows{} rows sampled, \TFfraud{} fraud): the synthetic
credit-card transaction dataset released with TabBERT~\cite{padhi2021tabular}.
The original release includes two datasets; we use the credit-card subset,
referred to as ``TabFormer'' throughout. It contains 1
continuous feature (transaction
amount), 5 low-cardinality categoricals (use chip, error type, hour, weekday,
month), and 5 high-cardinality identifiers (Merchant Category Code (MCC) 109 values, merchant
state 200 values, merchant name 1{,}397 values, merchant city 2{,}075 values,
ZIP code 3{,}109 values); one-hot encoding these columns is infeasible at 2M
rows (the $\sim$6{,}600-dimensional matrix exceeds available memory).
Categories appearing fewer than 100 times are collapsed to an \textsc{other}
sentinel, following standard practice.

\subsection{Baselines}
\begin{itemize}
  \item \textit{Raw}: z-score standardization of numeric features plus one-hot
    encoding of low-cardinality categoricals (high-cardinality columns excluded,
    as dense one-hot encoding is infeasible at this scale). This is a common
    practitioner default; other categorical encoders (target or hashing
    encoders) are not evaluated here.
  \item \textit{PCA-16}: Principal Component Analysis (PCA) applied to the raw matrix, reducing to 16
    dimensions, the classical linear dimensionality-reduction baseline at the
    same latent size as CARTA.
  \item \textit{CatBoost}~\cite{prokhorenkova2018catboost}: a gradient-boosted
    decision tree with native high-cardinality categorical handling
    (ordered target statistics), trained fully supervised on all
    features, including the merchant identifiers that Raw must drop. Unlike
    CARTA it requires task labels and is retrained per task; it is our strongest
    fully-supervised baseline (200 iterations, depth 6, balanced class weights,
    5~seeds, identical splits).
  \item \textit{CARTA-16}: frozen 16-dim latent means from the trained
    CARTA encoder.
\end{itemize}
GBDT denotes gradient-boosted decision trees~\cite{friedman2001greedy}, using
the histogram-based \texttt{scikit-learn}~\cite{pedregosa2011scikit}
implementation (max depth 6, 200 estimators). Results are means $\pm$ standard
deviations with stratified 70/30 train/test splits; each table states whether
the reported variation is over training seeds (independent encoder runs) or
probe seeds (re-fit probes on a fixed representation).

\subsection{Evaluation Protocol}
\label{sec:protocol}

All representations (raw, PCA-16, and CARTA-16) are evaluated through the
same two probes: (i)~a \textit{linear probe} (logistic regression,
class-weight balanced), measuring pure representation quality; and (ii)~a
\textit{GBDT probe}, measuring the upper-bound downstream utility.
All representations are evaluated with frozen parameters; no labels are used
during the representation learning phase of CARTA or PCA.

Our primary metric is the area under the precision--recall curve
(\textit{AUPRC}; equivalently average precision). We also report the area under
the ROC curve (ROC-AUC) for the GBDT probe. AUPRC is the appropriate metric for
highly imbalanced fraud detection, since ROC-AUC can be misleadingly high even
for weak classifiers~\cite{jesus2022turning}.

\textit{Why linear probes.}
Following standard SSL practice~\cite{oord2018representation}, the linear probe
measures \emph{representation quality}: a frozen embedding is good when a simple
linear classifier separates the classes from it. GBDT on raw features
(Table~\ref{tab:main}) is reported alongside as a strong fully-supervised
baseline.

\textit{Label-efficiency evaluation.}
To assess the practical value of self-supervised pretraining, each probe is
also trained on stratified subsets of 1\%, 5\%, 10\%, 25\%, and 100\% of
training labels, with the test set held fixed.

\textit{Implementation.}
All CARTA models are trained on a single NVIDIA A100 GPU using JAX/Flax,
batch size 16{,}384, Adam optimizer with warmup-cosine schedule
(peak $5\!\times\!10^{-4}$, 10 warmup epochs), up to 300 epochs with early
stopping (patience 50) on held-out self-supervised loss. Hyperparameters
($Q{=}8$, $L{=}4$, $z{=}16$, $p{=}0.2$, $\tau{=}0.07$, $\lambda_c{=}0.1$)
are standard defaults held fixed across datasets. Code and configuration files
are available at \url{https://github.com/Shreyaskc/carta}; both benchmarks are
publicly available.

% ─────────────────────────────────────────────────────────────────────────────
\section{Results}
\label{sec:results}

\subsection{Main Representation-Quality Results}

Table~\ref{tab:main} reports full-label probe performance.

\begin{table*}[t]
\centering
\caption{Full-label probe performance (mean $\pm$ std); Dim = representation
dimensionality. \textbf{Bold} marks the best frozen representation per column;
the fully-supervised CatBoost baseline ($\S$) is excluded. $\dagger$~Raw drops
the high-cardinality columns (OHE infeasible); $\ddagger$~frequency encoding
applies only to TabFormer. Subscript stds: CARTA-16 on IEEE-CIS is the mean
over 5 independent training seeds (population std; matched inner probe scores
averaged per run); Raw/PCA representations are deterministic, so their stds
reflect 5 probe seeds. TabFormer results are from a single training run
(5 probe seeds).}
\label{tab:main}
\small
\setlength{\tabcolsep}{6pt}
\begin{tabular}{@{}lrcccccc@{}}
\toprule
& & \multicolumn{3}{c}{\textit{IEEE-CIS} (\IEEErows{} rows, \IEEEfraud{} fraud)}
  & \multicolumn{3}{c}{\textit{TabFormer} (\TFrows{} rows, \TFfraud{} fraud)} \\
\cmidrule(lr){3-5}\cmidrule(lr){6-8}
Method & Dim & \makecell{Lin.\\AUPRC} & \makecell{GBDT\\AUPRC} & \makecell{GBDT\\AUC}
             & \makecell{Lin.\\AUPRC} & \makecell{GBDT\\AUPRC} & \makecell{GBDT\\AUC} \\
\midrule
Raw$^\dagger$        & 30/68 & \IEEErawLinAUPRC$_{\pm\IEEErawLinStd}$  & \best{\IEEErawGBDTAUPRC}$_{\pm\IEEErawGBDTStd}$ & \best{\IEEErawGBDTAUC}
                             & \TFrawLinAUPRC$_{\pm\TFrawLinStd}$   & \TFrawGBDTAUPRC$_{\pm\TFrawGBDTStd}$  & \TFrawGBDTAUC  \\
Freq.\ enc.$^\ddagger$ & 35/69 & ---             & ---                      & ---
                             & \best{\TFfreqLinAUPRC}$_{\pm\TFfreqLinStd}$  & \TFfreqGBDTAUPRC$_{\pm\TFfreqGBDTStd}$ & \TFfreqGBDTAUC \\
PCA-16               & 16    & \IEEEpcaLinAUPRC$_{\pm\IEEEpcaLinStd}$  & \IEEEpcaGBDTAUPRC$_{\pm\IEEEpcaGBDTStd}$        & \IEEEpcaGBDTAUC
                             & \TFpcaLinAUPRC$_{\pm\TFpcaLinStd}$   & \TFpcaGBDTAUPRC$_{\pm\TFpcaGBDTStd}$  & \TFpcaGBDTAUC  \\
\midrule
CARTA-16 (ours)      & 16    & \best{\IEEEgaterLinAUPRC}$_{\pm\IEEEgaterLinStd}$ & \IEEEgaterGBDTAUPRC$_{\pm\IEEEgaterGBDTStd}$ & \IEEEgaterGBDTAUC
                             & \TFgaterLinAUPRC$_{\pm\TFgaterLinStd}$   & \best{\TFgaterGBDTAUPRC}$_{\pm\TFgaterGBDTStd}$ & \best{\TFgaterGBDTAUC} \\
\midrule
CatBoost$^\S$        & all   & ---               & \IEEEcatboostGBDTAUPRC & \IEEEcatboostGBDTAUC
                             & ---               & \TFcatboostGBDTAUPRC & \TFcatboostGBDTAUC \\
\bottomrule
\end{tabular}
\end{table*}

\textit{IEEE-CIS.} On the linear probe, CARTA-16 ($\IEEEgaterLinAUPRC$)
outperforms raw features ($\IEEErawLinAUPRC$) despite using half the
dimensions: self-supervised pretraining yields a more linearly separable
representation than the raw feature space. On the GBDT probe, raw features
($\IEEErawGBDTAUPRC$) retain an advantage, as expected for a supervised GBDT
with full feature access; CARTA's value lies in label efficiency and multi-task
reuse (Section~\ref{sec:labeleff}).

\textit{TabFormer.}
The high-cardinality experiment illustrates two points. First, the practical
raw baseline, which must exclude merchant name (1,397 unique values), city
(2,075), and ZIP (3,109) because one-hot encoding them at 2M rows exceeds
available memory, achieves only GBDT AUPRC~$=\TFrawGBDTAUPRC$, far short of
the encodings that retain merchant identity. CARTA-16, encoding all merchant identifiers via
learned embedding tables, achieves GBDT AUPRC~$=\TFgaterGBDTAUPRC$, a
\TFgaterGBDTlift{} improvement over the raw baseline that excludes merchant
features entirely. A common alternative to one-hot encoding replaces
high-cardinality categories with log-normalized frequency
encoding~\cite{gorishniy2021revisiting}, which achieves
GBDT AUPRC~$=\TFfreqGBDTAUPRC$ at full 2M-row scale (Table~\ref{tab:main}).
CARTA ($\TFgaterGBDTAUPRC$) still outperforms this stronger baseline,
indicating that learned embedding tables capture richer merchant-identity
signal than frequency proxies. The merchant identifiers carry strong fraud
signal (fraud incidence is concentrated among subsets of merchants); among the
baselines evaluated here, CARTA is the only one that incorporates this signal
via learned embedding tables without requiring full label supervision.
The weak CARTA linear headline ($\TFgaterLinAUPRC$) reflects that at a
0.13\% base rate a class-weighted linear head over a 16-dim latent recovers
little; the merchant-identity signal in the embedding is nonlinear and is
extracted by the GBDT probe.
Note: the extreme class imbalance (0.13\% fraud, $\sim$760 test-set
positives) produces high variance; for the merchant-aware representations,
GBDT AUPRC standard deviations across the 5 probe seeds (re-split and re-fit
on a fixed embedding) range from 0.02 to 0.08.

\textit{Fully-supervised baseline (CatBoost).}
CatBoost, trained end-to-end on all features including native categoricals, is
the strongest baseline on TabFormer (GBDT AUPRC~$=\TFcatboostGBDTAUPRC$),
exceeding CARTA's frozen embedding ($\TFgaterGBDTAUPRC$). CatBoost consumes task
labels for this single task, while CARTA's embedding is label-free and reused
across tasks, so the two occupy different points on the supervision spectrum.
On IEEE-CIS (all low-cardinality), CatBoost ($\IEEEcatboostGBDTAUPRC$) sits below
raw-feature GBDT ($\IEEErawGBDTAUPRC$), reflecting hyperparameter defaults
(symmetric trees vs.\ depth-6 sklearn GBDT); this is consistent with prior
reports~\cite{prokhorenkova2018catboost}. The high-cardinality comparison is
therefore three-way: both learned embedding tables (CARTA, $\TFgaterGBDTAUPRC$)
and native-categorical boosting (CatBoost, $\TFcatboostGBDTAUPRC$) far
outperform the merchant-dropping raw baseline ($\TFrawGBDTAUPRC$) and the
frequency-encoding proxy ($\TFfreqGBDTAUPRC$). CatBoost attains the highest
single-task accuracy, and among the four methods CARTA is the one that yields a
\emph{label-free, reusable} representation.

\subsection{Label Efficiency}
\label{sec:labeleff}

\begin{figure}[t]
\centering
\includegraphics[width=\linewidth]{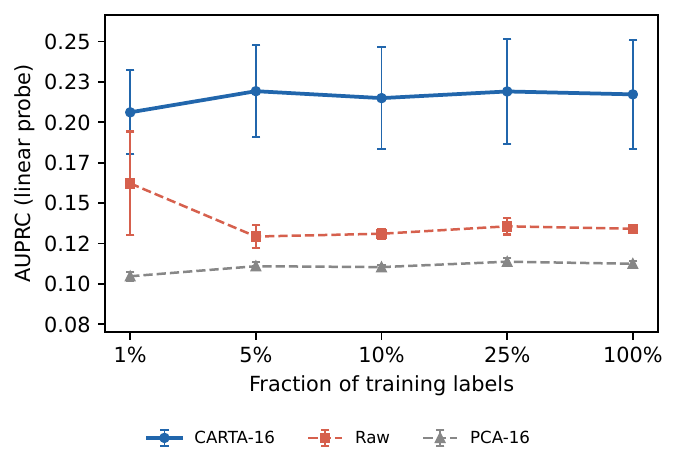}
\caption{Linear-probe AUPRC vs.\ fraction of training labels; error bars are
$\pm\sigma$ over 5 training seeds for CARTA-16 and 5 probe seeds for Raw and
PCA-16. On IEEE-CIS, CARTA-16 at 1\% of labels
($\IEEEgaterLEonepc$) exceeds raw features at 100\% of labels
($\IEEErawLinAUPRC$), so self-supervised pretraining reaches a given
representation quality with far fewer labels.}
\label{fig:labeleff}
\end{figure}

Figure~\ref{fig:labeleff} and Table~\ref{tab:labeleff} show linear-probe
AUPRC as the fraction of training labels is varied.

\begin{table}[t]
\centering
\caption{Linear-probe AUPRC on IEEE-CIS vs.\ fraction of training labels (mean;
5 training seeds for CARTA-16, 5 probe seeds otherwise). CARTA-16 at 1\% labels
exceeds raw features at 100\% labels; the
margin (CARTA-16 at 1\%~$=\IEEEgaterLEonepc$ vs.\ raw at 100\%~$=\IEEErawLinAUPRC$)
exceeds the full-label stds ($\pm\IEEEgaterLinStd$ over training seeds for
CARTA-16 and $\pm\IEEErawLinStd$ over probe seeds for raw, respectively).}
\label{tab:labeleff}
\footnotesize
\setlength{\tabcolsep}{4pt}
\begin{tabular}{@{}lccccc@{}}
\toprule
Method & 1\% & 5\% & 10\% & 25\% & 100\% \\
\midrule
Raw      & \IEEErawLEonepc & \IEEErawLEfivepc & \IEEErawLEtenpc & \IEEErawLEtwentyfivepc & \IEEErawLinAUPRC \\
PCA-16   & \IEEEpcaLEonepc & \IEEEpcaLEfivepc & \IEEEpcaLEtenpc & \IEEEpcaLEtwentyfivepc & \IEEEpcaLinAUPRC \\
UMAP-16  & \RCUMAPLEonepc & \RCUMAPLEfivepc & \RCUMAPLEtenpc & \RCUMAPLEtwentyfivepc & \RCUMAPLin \\
CARTA-16 & \best{\IEEEgaterLEonepc} & \best{\IEEEgaterLEfivepc} & \best{\IEEEgaterLEtenpc} & \best{\IEEEgaterLEtwentyfivepc} & \best{\IEEEgaterLinAUPRC} \\
\bottomrule
\end{tabular}
\end{table}

On IEEE-CIS, CARTA at 1\% of labels (AUPRC~$=\IEEEgaterLEonepc$) exceeds raw
features at 100\% ($\IEEErawLinAUPRC \pm \IEEErawLinStd$); at a matched 1\% budget
it also leads raw ($\IEEErawLEonepc$), so the advantage does not hinge on the
cross-fraction comparison. Both curves are mildly non-monotonic: at a 3.5\% base
rate a class-weighted linear head has high variance at small support and sits
near that base rate, so more labels need not monotonically help it. CARTA
front-loads the discriminative structure so few labels suffice, while PCA-16
reaches only $\IEEEpcaLEonepc$ at 1\%; the gain therefore comes from the learned
nonlinear compression, since a generic 16-dim projection does not achieve it.

\subsection{Ablation Study}
\label{sec:ablation}

Table~\ref{tab:ablation} compares training variants; the no-InfoNCE variant
also removes paired-view excision, so its effect cannot be attributed to the
contrastive loss alone.

\begin{table}[t]
\centering
\caption{Ablation results (linear probe / GBDT AUPRC; each variant is a
\emph{single} trained encoder, Perceiver unless noted, evaluated as the mean
over 5 probe seeds; up to 300 epochs with early stopping, batch 16{,}384
throughout). ``Full'' denotes the ablation reference configuration (cell-mode
excision + InfoNCE + cyclical KL). Rows are intended for within-table
comparison; the canonical CARTA numbers are those in Table~\ref{tab:main},
whose CARTA row instead averages 5 independent training runs with per-run
inner-probe averaging. The protocols therefore differ by seed population
(e.g., the feature-wise row here is one training run scoring 0.241, while the
Table~\ref{tab:main} 5-run mean is $0.217\pm0.034$). TabFormer values are
especially sensitive: the linear probe operates near the 0.13\% base rate, and
for the merchant-aware variants GBDT AUPRC varies by 0.02--0.08 (std) across
the 5 probe seeds even on a fixed embedding.}
\label{tab:ablation}
\footnotesize
\setlength{\tabcolsep}{4pt}
\resizebox{\linewidth}{!}{%
\begin{tabular}{@{}lcccc@{}}
\toprule
& \multicolumn{2}{c}{\textit{IEEE-CIS}} & \multicolumn{2}{c}{\textit{TabFormer}} \\
\cmidrule(lr){2-3}\cmidrule(lr){4-5}
Variant & Lin.\ AUPRC & GBDT AUPRC & Lin.\ AUPRC & GBDT AUPRC \\
\midrule
Full (cell + InfoNCE + cycl.\ KL)
  & \IEEEablFULLlin & \IEEEablFULLgbdt
  & \TFablFULLlin   & \TFablFULLgbdt \\
\midrule
$-$ Cyclical KL ($\beta{=}1$)
  & \IEEEablNOKLlin   & \IEEEablNOKLgbdt
  & \TFablNOKLlin     & \TFablNOKLgbdt \\
$-$ InfoNCE (single view)
  & \IEEEablNOCONTlin & \IEEEablNOCONTgbdt
  & \TFablNOCONTlin  & \TFablNOCONTgbdt \\
$-$ Token excision
  & \IEEEablNOEXClin  & \IEEEablNOEXCgbdt
  & \TFablNOEXClin   & \TFablNOEXCgbdt \\
Feature-wise excision (default)
  & \IEEEablFEATlin   & \IEEEablFEATgbdt
  & \TFablFEATlin    & \TFablFEATgbdt \\
\midrule
GRU encoder
  & \IEEEablGRUlin  & \IEEEablGRUgbdt
  & \TFablGRUlin    & \TFablGRUgbdt \\
MLP encoder (no attention)
  & \IEEEablMLPlin  & \IEEEablMLPgbdt
  & \TFablMLPlin    & \TFablMLPgbdt \\
\bottomrule
\end{tabular}}
\end{table}

\textit{InfoNCE contrastive objective.}
The no-InfoNCE variant shows the largest linear-probe drop on both datasets:
on IEEE-CIS from $\IEEEablFULLlin$ to $\IEEEablNOCONTlin$ ($-0.048$ AUPRC);
on TabFormer from $\TFablFULLlin$ to $\TFablNOCONTlin$ (about a 3$\times$
relative decrease). This variant trains on a single uncorrupted view, so the
drop measures the combined effect of the contrastive loss and its paired-view
excision augmentation, not the loss in isolation.

\textit{Cyclical KL annealing.}
Replacing the cyclical schedule with a fixed $\beta{=}1$ (no annealing) slightly
improves IEEE-CIS linear AUPRC ($\IEEEablNOKLlin$ vs.\ $\IEEEablFULLlin$) and
markedly improves TabFormer ($\TFablNOKLlin$ vs.\ $\TFablFULLlin$). Even in the
label-efficient regime the two are tied on IEEE-CIS (1\%-label linear AUPRC
$\IEEEablFULLLEonepc$ with annealing vs.\ $\IEEEablNOKLLEonepc$ without). We
therefore make no representation-quality claim for cyclical annealing: its
documented role is to stabilize VAE training against posterior
collapse~\cite{fu2019cyclical}, and we keep it in the Full configuration for
collapse prevention and one consistent recipe across datasets, not for any
downstream gain. The large TabFormer gap, and whether the no-annealing gains
survive retuning, warrant a dedicated schedule study we leave to future work.

\textit{Token excision.}
The ``$-$~Token excision'' row sets $p{=}0$ while keeping InfoNCE; the two
views then differ only via VAE reparameterisation and dropout, giving a weak
contrastive signal (linear probe: $\IEEEablNOEXClin$ / $\TFablNOEXClin$).

\textit{Excision mode and SCARF: a controlled 5-seed comparison.}
We compare feature-wise and cell-mode token excision against SCARF-style marginal
shuffling~\cite{bahri2022scarf}, under an identical Perceiver encoder and budget
and each trained over 5 independent seeds (IEEE-CIS only; training-seed std). On
IEEE-CIS the three linear-probe means are feature-wise
$\IEEEexcFeatLin{\pm}\IEEEexcFeatLinStd$, cell-mode
$\IEEEexcCellLin{\pm}\IEEEexcCellLinStd$, and SCARF
$\IEEEexcScarfLin{\pm}\IEEEexcScarfLinStd$: a spread comparable to the per-mode
training-seed standard deviations, so with 5 seeds per mode we observe no clear
separation and conclude the
excision granularity does \emph{not} drive quality on this low-embedding dataset.
The choice does matter on embedding-heavy data. Cell-mode never masks the learned
embedding tokens, so on TabFormer it corrupts only the continuous and
low-cardinality tokens, whereas feature-wise excision also masks whole embedding
features and attains the strongest TabFormer high-cardinality GBDT (Table~\ref{tab:main}). We therefore
adopt feature-wise excision as the default, not for a quality edge on IEEE, but
because it is what makes token excision act on high-cardinality embedding features.
An earlier single-seed reading had suggested a broader feature-wise advantage on
IEEE; that did not survive multi-seed evaluation, which we report for transparency.

\textit{Encoder architecture.}
All three encoder architectures (Perceiver, GRU-gated, MLP-VAE) are
competitive. The GRU encoder leads on TabFormer GBDT AUPRC
($\TFablGRUgbdt$, vs.\ MLP $\TFablMLPgbdt$ and Perceiver $\TFablFULLgbdt$)
and on IEEE-CIS linear AUPRC ($\IEEEablGRUlin$, vs.\ MLP $\IEEEablMLPlin$
and Perceiver $\IEEEablFULLlin$). At $N{=}7$ tokens, the GRU's iterative
context refinement over a short token sequence is as effective as maintaining
$Q{=}8$ parallel latent queries; the MLP is also competitive on IEEE-CIS GBDT
($\IEEEablMLPgbdt$). All three architectures share the same tokenization,
token excision, and training objective, so the differences reflect encoder
capacity rather than any other factor. The Perceiver's O($Q{\cdot}N$) decoupled
complexity becomes advantageous only as feature count grows toward the 30+
heterogeneous columns of full industrial card-network datasets; at small $N$
the simpler baselines are credible alternatives. We thus present the Perceiver
as the most scalable of the three, competitive at small $N$ and increasingly
advantageous as the feature-token count grows.

% ─────────────────────────────────────────────────────────────────────────────
\subsection{Comparison to Unsupervised Representation Baselines}
\label{sec:repcomp}

\begin{figure}[t]
\centering
\includegraphics[width=\linewidth]{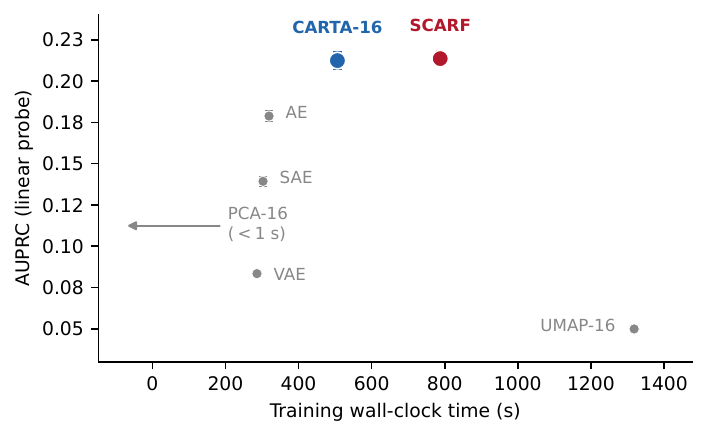}
\caption{Representation quality (linear-probe AUPRC, 5 probe seeds) versus
training wall-clock time for seven label-free 16-dim encoders on IEEE-CIS
(one timed training run per method). CARTA's (blue) full-label linear AUPRC is
similar to SCARF's at \RCgaterTimeFracScarf{} of SCARF's single-run training
time. Error bars are $\pm\sigma$ over probe seeds.}
\label{fig:repcomp}
\end{figure}

Tables~\ref{tab:main}--\ref{tab:ablation} compare CARTA against supervised
models (raw-feature GBDT, CatBoost); a representation, though, is best judged
against other representations. We therefore compare CARTA to seven label-free
encoders that all produce a 16-dim embedding from the same inputs, scoring each
with the same linear and GBDT probes, K-means silhouette, and training-time
budget (Table~\ref{tab:repcomp}, Figure~\ref{fig:repcomp}): PCA and
Uniform Manifold Approximation and Projection (UMAP)~\cite{mcinnes2018umap} (non-linear dimensionality reduction), a
plain autoencoder (AE), a sparse autoencoder (SAE), a vanilla MLP-VAE (no excision
or contrastive objective), and SCARF~\cite{bahri2022scarf}. Supervised GBDT and
CatBoost stay a separate upper reference (Section~\ref{sec:discussion}).

\begin{table}[t]
\centering
\caption{Unsupervised representation comparison on IEEE-CIS. Every method
produces a 16-dim label-free embedding scored with the same probes, K-means
silhouette (best $k\in[3,7]$), and training wall-time. Lin.\ AUPRC is the mean
over 5 probe seeds; LE@1\% is that probe at 1\% of training labels; each
embedding is a single timed training run (UMAP timed
multi-threaded). In this single-run comparison CARTA's full-label linear AUPRC
($\RCGATERLin$) is similar to SCARF's ($\RCSCARFLin$) with a shorter wall-clock
time (\RCgaterTimeFracScarf{} of SCARF's), while SCARF leads on LE@1\%
($\RCSCARFLEonepc$ vs.\ $\RCGATERLEonepc$) and GBDT AUPRC ($\RCSCARFGbdt$ vs.\
$\RCGATERGbdt$); the high-silhouette methods (MLP-VAE, UMAP) have weak probes,
so silhouette does not track downstream usefulness here.}
\label{tab:repcomp}
\footnotesize
\setlength{\tabcolsep}{4pt}
\begin{tabular}{@{}lccccc@{}}
\toprule
Representation & Lin.\ AUPRC & LE@1\% & GBDT AUPRC & Silhou. & Train (s) \\
\midrule
PCA-16    & \RCPCALin   & \RCPCALEonepc   & \RCPCAGbdt   & \RCPCASil   & \RCPCATrain \\
UMAP-16   & \RCUMAPLin  & \RCUMAPLEonepc  & \RCUMAPGbdt  & \RCUMAPSil  & \RCUMAPTrain \\
Autoencoder & \RCAELin  & \RCAELEonepc    & \RCAEGbdt    & \RCAESil    & \RCAETrain \\
Sparse AE & \RCSAELin   & \RCSAELEonepc   & \RCSAEGbdt   & \RCSAESil   & \RCSAETrain \\
MLP-VAE   & \RCVVAELin  & \RCVVAELEonepc  & \RCVVAEGbdt  & \RCVVAESil  & \RCVVAETrain \\
SCARF     & \RCSCARFLin & \RCSCARFLEonepc & \RCSCARFGbdt & \RCSCARFSil & \RCSCARFTrain \\
\textbf{CARTA-16 (ours)} & \RCGATERLin & \RCGATERLEonepc & \RCGATERGbdt & \RCGATERSil & \RCGATERTrain \\
\bottomrule
\end{tabular}
\end{table}

CARTA and SCARF lead on the full-label linear probe, and both encode
non-linear structure.
Among label-free encoders, CARTA and SCARF attain the highest linear-probe
AUPRC ($\RCGATERLin$ and $\RCSCARFLin$), roughly \RCgaterPcaLinRatio{} the score
of a PCA projection ($\RCPCALin$) and well above the autoencoder ($\RCAELin$),
sparse autoencoder ($\RCSAELin$), and MLP-VAE ($\RCVVAELin$). Non-linearity alone
does not explain the gap: UMAP-16, a non-linear manifold embedding of the same
inputs, is the \emph{weakest} representation ($\RCUMAPLin$, below even linear
PCA), because it optimizes local neighborhood structure for visualization rather
than linear separability of a downstream signal. The advantage therefore comes
from the self-supervised objective, not from non-linearity per se.

CARTA reached this full-label linear-probe quality with less compute in our
single timed run: $\RCGATERTrain$~s versus $\RCSCARFTrain$~s for SCARF (which
took \RCscarfTimeExtra{} longer), plausibly because token excision is cheaper
than SCARF's marginal shuffling, which resamples each corrupted cell from the
empirical column distribution. These are one timing per method on one machine,
not a general speed claim, and SCARF leads on LE@1\% ($\RCSCARFLEonepc$ vs.\
$\RCGATERLEonepc$) and the GBDT probe ($\RCSCARFGbdt$ vs.\ $\RCGATERGbdt$) in
the same comparison. The controlled multi-seed excision-vs-SCARF quality
comparison is reported in Section~\ref{sec:ablation}.

Silhouette does not track representation quality under extreme imbalance.
The MLP-VAE attains by far the highest silhouette ($\RCVVAESil$) yet a weak
linear probe ($\RCVVAELin$), and UMAP-16 pairs a moderate silhouette
($\RCUMAPSil$) with the lowest probe of all ($\RCUMAPLin$), while CARTA and SCARF
have the lowest silhouettes ($\RCGATERSil$, $\RCSCARFSil$) yet the strongest
probes. Cluster compactness is thus a poor proxy for fraud-relevant structure at
a 3.5\% base rate, so we judge representation quality with supervised probes
rather than a projection visual.

% ─────────────────────────────────────────────────────────────────────────────
\subsection{Latent Space Structure and Unsupervised Risk Stratification}
\label{sec:latent}

\begin{figure}[t]
\centering
\includegraphics[width=\linewidth]{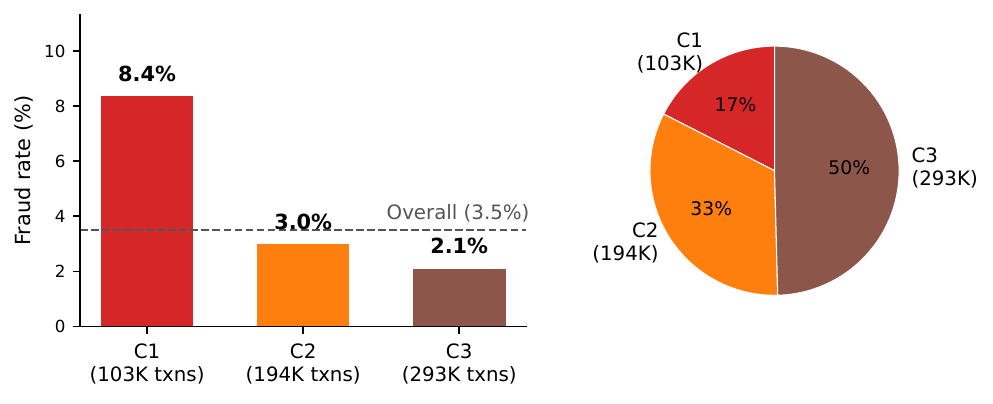}
\caption{Unsupervised behavioral archetypes discovered by K-means
($k=\EPkbest$, silhouette $=\EPsil$) on the full IEEE-CIS CARTA latent space
(590K transactions, no fraud labels used). The highest-incidence archetype
(\EPConesize{} transactions) shows an observed fraud rate of \EPConefraud{},
nearly \EPfraudratio{} the rate of the lowest-incidence archetype
(\EPCthreefraud{}), despite being discovered without any labeled supervision.}
\label{fig:entity}
\end{figure}

Figure~\ref{fig:entity} shows unsupervised risk stratification on IEEE-CIS:
K-means on the 16-dim latent (no fraud labels) recovers \EPkbest{} behavioral
archetypes with a \EPfraudratio{} fraud-rate spread, and the clustering is stable
(pairwise ARI $=\ClusterARI \pm \ClusterARIstd$ over 10 seeds). As in
Section~\ref{sec:repcomp}, the low silhouette ($\EPsil$) reflects archetype
overlap that nonetheless differs sharply in fraud rate; the spread carries the
discriminative signal that could support analyst triage.

The same clustering on the 2M-transaction TabFormer latent ($k=4$, silhouette
$=0.141$) recovers merchant-type archetypes with a factor of more than 20
variation in observed fraud incidence across groups (from 0.01\% to 0.21\%),
against a 0.127\% overall rate. Since merchant identity is itself an encoder
input, recovering merchant-type groupings is expected by construction; the
fraud-incidence spread across groups is not directly supervised and may partly
reflect transaction frequency or merchant mix, so we treat these clusters as
exploratory groupings rather than causal fraud signals.

% ─────────────────────────────────────────────────────────────────────────────
\section{Discussion and Limitations}
\label{sec:discussion}

\textit{Fully-supervised gradient boosting is stronger on peak accuracy.}
On IEEE-CIS, raw-feature GBDT ($\IEEErawGBDTAUPRC$) leads; on TabFormer, CatBoost
with native categoricals ($\TFcatboostGBDTAUPRC$) is strongest. CARTA does not
beat these fully-supervised, single-task models on peak accuracy and we do not
claim it does (CatBoost is itself label-efficient, IEEE 1\%-label AUPRC
$\IEEEcatboostLEonepc$). Its contribution is complementary: a single
\emph{label-free} embedding reused across the analytical stack with no per-task
labels or retraining. With abundant labels for one fixed task, CatBoost is the
better choice; when one consistent representation is needed across many evolving
tasks, CARTA is the setting we target.

\textit{Class imbalance and AUPRC variance.}
TabFormer's 0.13\% fraud rate leaves $\sim$760 test positives, so AUPRC is
sensitive to seed noise (GBDT AUPRC std 0.02--0.08 over 5 probe seeds for the
merchant-aware representations). A \TFgaterGBDTlift{} improvement over raw is
robust, though per-variant TabFormer differences should be read as indicative:
every TabFormer number comes from a single trained encoder per variant, and
replication over independent training seeds was performed only on IEEE-CIS, so
a larger sweep on this imbalanced benchmark is warranted.

\textit{Evaluation splits and potential leakage.}
Our primary splits are random 70/30. Because the original IEEE-CIS split is
temporal, random splits are optimistic for time-ordered fraud; learned merchant
embeddings also permit \emph{entity leakage} when the same merchants appear in
both train and test. All representations are fit on the pooled feature matrix
without labels and each probe is trained only on the training split, so the
transductive feature setting is uniform across methods and no labels enter
representation learning. We also evaluate under a \emph{temporal} split (last 30\% by date as test): the
linear-probe advantage persists (CARTA 0.240 vs.\ raw 0.180; the matched
single-run random split gives 0.198/0.134), so the representation-quality
finding is not a random-split artifact.
Merchant-disjoint TabFormer splits remain as future work.

\textit{Few tokens on the evaluated datasets.}
Both benchmarks are small ($N{=}2$ on IEEE-CIS, $N{=}7$ on TabFormer), so the
Perceiver's O($Q{\cdot}N$) advantage is not exercised and an MLP encoder is
competitive (Table~\ref{tab:ablation}). The architecture targets the 30+
heterogeneous-column regime of production data; validation there is future work.

\textit{Cold start.}
Unseen merchant identifiers map to a reserved $\langle\text{unknown}\rangle$
embedding index at inference, distinct from the \textsc{other} bucket that
collapses rare categories during vocabulary fitting; newly appearing
merchants can be the riskiest, so the embedding advantage is smallest where it
could matter most. Hashing or side-information fallbacks are future work.

\textit{Scope and comparison breadth.}
We evaluate the frozen embedding on two downstream tasks (label-efficient fraud
probing and unsupervised risk stratification); drift monitoring and forecasting
remain motivating uses we do not yet evaluate. An end-to-end
FT-Transformer~\cite{gorishniy2021revisiting} on IEEE-CIS reaches AUPRC $\FTTonepc$
at 1\% and $\FTTfull$ at 100\% of labels but is supervised and per-task (no
reusable embedding). Table~\ref{tab:repcomp} covers comparable SSL methods (SCARF,
AE, SAE); a full comparison to VIME and SubTab is a natural follow-on. The latent
width $z{=}16$ was chosen to match PCA-16 and not tuned; a sensitivity study is
straightforward with the released code. All results are reported on public research
benchmarks and reflect research findings, not organizational policy or operational
practice.

% ─────────────────────────────────────────────────────────────────────────────
\section{Conclusion}
\label{sec:conclusion}

We presented CARTA, a self-supervised encoder for compact, reusable embeddings
from raw heterogeneous payment transaction data. On two public benchmarks,
CARTA achieves label efficiency (1\% of labels exceeds a raw-feature probe at
100\%) and high-cardinality capability (outperforming baselines that must drop
merchant identifiers and frequency-encoding proxies that replace them), while
fully-supervised CatBoost remains stronger on peak single-task accuracy. In the
ablation, removing the contrastive objective together with its paired-view
augmentation (a compound intervention) is the most damaging change; the spread
across token-excision modes on IEEE-CIS is comparable to per-mode training-seed
variability, with feature-wise excision retained to
extend corruption to embedding features. Unsupervised K-means on the frozen latent recovers
behavioral archetypes whose observed fraud incidence differs markedly across
groups without label supervision. The encoder is trained once and reused across
downstream tasks; evaluation on drift monitoring and forecasting remains future
work.

\bibliographystyle{ACM-Reference-Format}
\bibliography{references}

%%% -*-BibTeX-*-
%%% Do NOT edit. File created by BibTeX with style
%%% ACM-Reference-Format-Journals [18-Jan-2012].

\begin{thebibliography}{26}

%%% ====================================================================
%%% NOTE TO THE USER: you can override these defaults by providing
%%% customized versions of any of these macros before the \bibliography
%%% command.  Each of them MUST provide its own final punctuation,
%%% except for \shownote{} and \showURL{}.  The latter two
%%% do not use final punctuation, in order to avoid confusing it with
%%% the Web address.
%%%
%%% To suppress output of a particular field, define its macro to expand
%%% to an empty string, or better, \unskip, like this:
%%%
%%% \newcommand{\showURL}[1]{\unskip}   % LaTeX syntax
%%%
%%% \def \showURL #1{\unskip}           % plain TeX syntax
%%%
%%% ====================================================================

\ifx \showCODEN    \undefined \def \showCODEN     #1{\unskip}     \fi
\ifx \showISBNx    \undefined \def \showISBNx     #1{\unskip}     \fi
\ifx \showISBNxiii \undefined \def \showISBNxiii  #1{\unskip}     \fi
\ifx \showISSN     \undefined \def \showISSN      #1{\unskip}     \fi
\ifx \showLCCN     \undefined \def \showLCCN      #1{\unskip}     \fi
\ifx \shownote     \undefined \def \shownote      #1{#1}          \fi
\ifx \showarticletitle \undefined \def \showarticletitle #1{#1}   \fi
\ifx \showURL      \undefined \def \showURL       {\relax}        \fi
% The following commands are used for tagged output and should be
% invisible to TeX
\providecommand\bibfield[2]{#2}
\providecommand\bibinfo[2]{#2}
\providecommand\natexlab[1]{#1}
\providecommand\showeprint[2][]{arXiv:#2}

\bibitem[Arik and Pfister(2021)]%
        {arik2021tabnet}
\bibfield{author}{\bibinfo{person}{Sercan~{\"O}. Arik} {and}
  \bibinfo{person}{Tomas Pfister}.} \bibinfo{year}{2021}\natexlab{}.
\newblock \showarticletitle{{TabNet}: Attentive Interpretable Tabular
  Learning}. In \bibinfo{booktitle}{\emph{Proceedings of the 35th AAAI
  Conference on Artificial Intelligence (AAAI)}}. \bibinfo{publisher}{AAAI
  Press}, \bibinfo{pages}{6679--6687}.
\newblock
\href{https://doi.org/10.1609/aaai.v35i8.16826}{doi:\nolinkurl{10.1609/aaai.v35i8.16826}}


\bibitem[Bahri et~al\mbox{.}(2022)]%
        {bahri2022scarf}
\bibfield{author}{\bibinfo{person}{Dara Bahri}, \bibinfo{person}{Heinrich
  Jiang}, \bibinfo{person}{Yi Tay}, {and} \bibinfo{person}{Donald Metzler}.}
  \bibinfo{year}{2022}\natexlab{}.
\newblock \showarticletitle{{SCARF}: Self-Supervised Contrastive Learning using
  Random Feature Corruption}. In \bibinfo{booktitle}{\emph{Proceedings of the
  10th International Conference on Learning Representations (ICLR)}}.
  \bibinfo{publisher}{OpenReview.net}.
\newblock


\bibitem[Cho et~al\mbox{.}(2014)]%
        {cho2014learning}
\bibfield{author}{\bibinfo{person}{Kyunghyun Cho}, \bibinfo{person}{Bart van
  Merri{\"e}nboer}, \bibinfo{person}{Caglar Gulcehre}, \bibinfo{person}{Dzmitry
  Bahdanau}, \bibinfo{person}{Fethi Bougares}, \bibinfo{person}{Holger
  Schwenk}, {and} \bibinfo{person}{Yoshua Bengio}.}
  \bibinfo{year}{2014}\natexlab{}.
\newblock \showarticletitle{Learning Phrase Representations using {RNN}
  Encoder--Decoder for Statistical Machine Translation}. In
  \bibinfo{booktitle}{\emph{Proceedings of the 2014 Conference on Empirical
  Methods in Natural Language Processing (EMNLP)}}.
  \bibinfo{publisher}{Association for Computational Linguistics},
  \bibinfo{pages}{1724--1734}.
\newblock
\href{https://doi.org/10.3115/v1/D14-1179}{doi:\nolinkurl{10.3115/v1/D14-1179}}


\bibitem[Clauset et~al\mbox{.}(2009)]%
        {clauset2009power}
\bibfield{author}{\bibinfo{person}{Aaron Clauset},
  \bibinfo{person}{Cosma~Rohilla Shalizi}, {and} \bibinfo{person}{Mark E.~J.
  Newman}.} \bibinfo{year}{2009}\natexlab{}.
\newblock \showarticletitle{Power-Law Distributions in Empirical Data}.
\newblock \bibinfo{journal}{\emph{SIAM Rev.}} \bibinfo{volume}{51},
  \bibinfo{number}{4} (\bibinfo{year}{2009}), \bibinfo{pages}{661--703}.
\newblock
\href{https://doi.org/10.1137/070710111}{doi:\nolinkurl{10.1137/070710111}}


\bibitem[Friedman(2001)]%
        {friedman2001greedy}
\bibfield{author}{\bibinfo{person}{Jerome~H. Friedman}.}
  \bibinfo{year}{2001}\natexlab{}.
\newblock \showarticletitle{Greedy Function Approximation: A Gradient Boosting
  Machine}.
\newblock \bibinfo{journal}{\emph{The Annals of Statistics}}
  \bibinfo{volume}{29}, \bibinfo{number}{5} (\bibinfo{year}{2001}),
  \bibinfo{pages}{1189--1232}.
\newblock
\href{https://doi.org/10.1214/aos/1013203451}{doi:\nolinkurl{10.1214/aos/1013203451}}


\bibitem[Fu et~al\mbox{.}(2019)]%
        {fu2019cyclical}
\bibfield{author}{\bibinfo{person}{Hao Fu}, \bibinfo{person}{Chunyuan Li},
  \bibinfo{person}{Xiaodong Liu}, \bibinfo{person}{Jianfeng Gao},
  \bibinfo{person}{Asli Celikyilmaz}, {and} \bibinfo{person}{Lawrence Carin}.}
  \bibinfo{year}{2019}\natexlab{}.
\newblock \showarticletitle{Cyclical Annealing Schedule: A Simple Approach to
  Mitigating {KL} Vanishing}. In \bibinfo{booktitle}{\emph{Proceedings of the
  2019 Conference of the North American Chapter of the Association for
  Computational Linguistics: Human Language Technologies (NAACL-HLT)}}.
  \bibinfo{publisher}{Association for Computational Linguistics},
  \bibinfo{pages}{240--250}.
\newblock
\href{https://doi.org/10.18653/v1/N19-1021}{doi:\nolinkurl{10.18653/v1/N19-1021}}


\bibitem[Gorishniy et~al\mbox{.}(2021)]%
        {gorishniy2021revisiting}
\bibfield{author}{\bibinfo{person}{Yury Gorishniy}, \bibinfo{person}{Ivan
  Rubachev}, \bibinfo{person}{Valentin Khrulkov}, {and} \bibinfo{person}{Artem
  Babenko}.} \bibinfo{year}{2021}\natexlab{}.
\newblock \showarticletitle{Revisiting Deep Learning Models for Tabular Data}.
  In \bibinfo{booktitle}{\emph{Advances in Neural Information Processing
  Systems 34 (NeurIPS)}}. \bibinfo{publisher}{Curran Associates, Inc.},
  \bibinfo{pages}{18932--18943}.
\newblock


\bibitem[Guo and Berkhahn(2016)]%
        {guo2016entity}
\bibfield{author}{\bibinfo{person}{Cheng Guo} {and} \bibinfo{person}{Felix
  Berkhahn}.} \bibinfo{year}{2016}\natexlab{}.
\newblock \showarticletitle{Entity Embeddings of Categorical Variables}.
\newblock \bibinfo{journal}{\emph{arXiv preprint arXiv:1604.06737}}
  (\bibinfo{year}{2016}).
\newblock


\bibitem[Hollmann et~al\mbox{.}(2025)]%
        {hollmann2025tabpfn}
\bibfield{author}{\bibinfo{person}{Noah Hollmann}, \bibinfo{person}{Samuel
  M{\"u}ller}, \bibinfo{person}{Lennart Purucker}, \bibinfo{person}{Arjun
  Krishnakumar}, \bibinfo{person}{Max K{\"o}rfer}, \bibinfo{person}{Shi~Bin
  Hoo}, \bibinfo{person}{Robin~Tibor Schirrmeister}, {and}
  \bibinfo{person}{Frank Hutter}.} \bibinfo{year}{2025}\natexlab{}.
\newblock \showarticletitle{Accurate predictions on small data with a tabular
  foundation model}.
\newblock \bibinfo{journal}{\emph{Nature}}  \bibinfo{volume}{637}
  (\bibinfo{year}{2025}), \bibinfo{pages}{319--326}.
\newblock
\href{https://doi.org/10.1038/s41586-024-08328-6}{doi:\nolinkurl{10.1038/s41586-024-08328-6}}


\bibitem[Huang et~al\mbox{.}(2020)]%
        {huang2020tabtransformer}
\bibfield{author}{\bibinfo{person}{Xin Huang}, \bibinfo{person}{Ashish Khetan},
  \bibinfo{person}{Milan Cvitkovic}, {and} \bibinfo{person}{Zohar Karnin}.}
  \bibinfo{year}{2020}\natexlab{}.
\newblock \showarticletitle{{TabTransformer}: Tabular Data Modeling using
  Contextual Embeddings}.
\newblock \bibinfo{journal}{\emph{arXiv preprint arXiv:2012.06678}}
  (\bibinfo{year}{2020}).
\newblock


\bibitem[{IEEE Computational Intelligence Society} and {Vesta
  Corporation}(2019)]%
        {ieeecis2019}
\bibfield{author}{\bibinfo{person}{{IEEE Computational Intelligence Society}}
  {and} \bibinfo{person}{{Vesta Corporation}}.}
  \bibinfo{year}{2019}\natexlab{}.
\newblock \bibinfo{title}{{IEEE-CIS} Fraud Detection}.
\newblock \bibinfo{howpublished}{Kaggle competition}.
\newblock
\shownote{\url{https://www.kaggle.com/c/ieee-fraud-detection}}.
\newblock


\bibitem[Jaegle et~al\mbox{.}(2021)]%
        {jaegle2021perceiver}
\bibfield{author}{\bibinfo{person}{Andrew Jaegle}, \bibinfo{person}{Felix
  Gimeno}, \bibinfo{person}{Andy Brock}, \bibinfo{person}{Oriol Vinyals},
  \bibinfo{person}{Andrew Zisserman}, {and} \bibinfo{person}{Jo{\~a}o
  Carreira}.} \bibinfo{year}{2021}\natexlab{}.
\newblock \showarticletitle{Perceiver: General Perception with Iterative
  Attention}. In \bibinfo{booktitle}{\emph{Proceedings of the 38th
  International Conference on Machine Learning (ICML)}}
  \emph{(\bibinfo{series}{Proceedings of Machine Learning Research},
  Vol.~\bibinfo{volume}{139})}. \bibinfo{publisher}{PMLR},
  \bibinfo{pages}{4651--4664}.
\newblock


\bibitem[Jesus et~al\mbox{.}(2022)]%
        {jesus2022turning}
\bibfield{author}{\bibinfo{person}{S{\'e}rgio Jesus}, \bibinfo{person}{Jos{\'e}
  Pombal}, \bibinfo{person}{Duarte Alves}, \bibinfo{person}{Andr{\'e} Cruz},
  \bibinfo{person}{Pedro Saleiro}, \bibinfo{person}{Rita~P. Ribeiro},
  \bibinfo{person}{Jo{\~a}o Gama}, {and} \bibinfo{person}{Pedro Bizarro}.}
  \bibinfo{year}{2022}\natexlab{}.
\newblock \showarticletitle{Turning the Tables: Biased, Imbalanced, Dynamic
  Tabular Datasets for {ML} Evaluation}. In \bibinfo{booktitle}{\emph{Advances
  in Neural Information Processing Systems 35 (NeurIPS), Datasets and
  Benchmarks Track}}. \bibinfo{publisher}{Curran Associates, Inc.}
\newblock


\bibitem[Kim et~al\mbox{.}(2024)]%
        {kim2024carte}
\bibfield{author}{\bibinfo{person}{Myung~Jun Kim}, \bibinfo{person}{L{\'e}o
  Grinsztajn}, {and} \bibinfo{person}{Ga{\"e}l Varoquaux}.}
  \bibinfo{year}{2024}\natexlab{}.
\newblock \showarticletitle{{CARTE}: Pretraining and Transfer for Tabular
  Learning}. In \bibinfo{booktitle}{\emph{Proceedings of the 41st International
  Conference on Machine Learning (ICML)}}.
\newblock


\bibitem[Kingma and Welling(2014)]%
        {kingma2013auto}
\bibfield{author}{\bibinfo{person}{Diederik~P. Kingma} {and}
  \bibinfo{person}{Max Welling}.} \bibinfo{year}{2014}\natexlab{}.
\newblock \showarticletitle{Auto-Encoding Variational {Bayes}}. In
  \bibinfo{booktitle}{\emph{Proceedings of the 2nd International Conference on
  Learning Representations (ICLR)}}.
\newblock


\bibitem[Krutikov et~al\mbox{.}(2024)]%
        {fu2024challenging}
\bibfield{author}{\bibinfo{person}{Sergei Krutikov}, \bibinfo{person}{Bulat
  Khaertdinov}, \bibinfo{person}{Rodion Kiriukhin}, \bibinfo{person}{Shubham
  Agrawal}, {and} \bibinfo{person}{Kees~Jan de Vries}.}
  \bibinfo{year}{2024}\natexlab{}.
\newblock \showarticletitle{Challenging Gradient Boosted Decision Trees with
  Tabular Transformers for Fraud Detection at {Booking.com}}.
\newblock \bibinfo{journal}{\emph{arXiv preprint arXiv:2405.13692}}
  (\bibinfo{year}{2024}).
\newblock


\bibitem[McInnes et~al\mbox{.}(2018)]%
        {mcinnes2018umap}
\bibfield{author}{\bibinfo{person}{Leland McInnes}, \bibinfo{person}{John
  Healy}, {and} \bibinfo{person}{James Melville}.}
  \bibinfo{year}{2018}\natexlab{}.
\newblock \showarticletitle{{UMAP}: Uniform Manifold Approximation and
  Projection}.
\newblock \bibinfo{journal}{\emph{arXiv:1802.03426}} (\bibinfo{year}{2018}).
\newblock


\bibitem[Padhi et~al\mbox{.}(2021)]%
        {padhi2021tabular}
\bibfield{author}{\bibinfo{person}{Inkit Padhi}, \bibinfo{person}{Yair Schiff},
  \bibinfo{person}{Igor Melnyk}, \bibinfo{person}{Mattia Rigotti},
  \bibinfo{person}{Youssef Mroueh}, \bibinfo{person}{Pierre Dognin},
  \bibinfo{person}{Jerret Ross}, \bibinfo{person}{Ravi Nair}, {and}
  \bibinfo{person}{Erik Altman}.} \bibinfo{year}{2021}\natexlab{}.
\newblock \showarticletitle{Tabular Transformers for Modeling Multivariate Time
  Series}. In \bibinfo{booktitle}{\emph{Proceedings of the 2021 IEEE
  International Conference on Acoustics, Speech and Signal Processing
  (ICASSP)}}. \bibinfo{publisher}{IEEE}, \bibinfo{pages}{3565--3569}.
\newblock
\href{https://doi.org/10.1109/ICASSP39728.2021.9414142}{doi:\nolinkurl{10.1109/ICASSP39728.2021.9414142}}


\bibitem[Pedregosa et~al\mbox{.}(2011)]%
        {pedregosa2011scikit}
\bibfield{author}{\bibinfo{person}{Fabian Pedregosa}, \bibinfo{person}{Ga{\"e}l
  Varoquaux}, \bibinfo{person}{Alexandre Gramfort}, \bibinfo{person}{Vincent
  Michel}, \bibinfo{person}{Bertrand Thirion}, \bibinfo{person}{Olivier
  Grisel}, \bibinfo{person}{Mathieu Blondel}, \bibinfo{person}{Peter
  Prettenhofer}, \bibinfo{person}{Ron Weiss}, \bibinfo{person}{Vincent
  Dubourg}, \bibinfo{person}{Jake Vanderplas}, \bibinfo{person}{Alexandre
  Passos}, \bibinfo{person}{David Cournapeau}, \bibinfo{person}{Matthieu
  Brucher}, \bibinfo{person}{Matthieu Perrot}, {and}
  \bibinfo{person}{{\'E}douard Duchesnay}.} \bibinfo{year}{2011}\natexlab{}.
\newblock \showarticletitle{Scikit-learn: Machine Learning in {Python}}.
\newblock \bibinfo{journal}{\emph{Journal of Machine Learning Research}}
  \bibinfo{volume}{12} (\bibinfo{year}{2011}), \bibinfo{pages}{2825--2830}.
\newblock


\bibitem[Prokhorenkova et~al\mbox{.}(2018)]%
        {prokhorenkova2018catboost}
\bibfield{author}{\bibinfo{person}{Liudmila Prokhorenkova},
  \bibinfo{person}{Gleb Gusev}, \bibinfo{person}{Aleksandr Vorobev},
  \bibinfo{person}{Anna~Veronika Dorogush}, {and} \bibinfo{person}{Andrey
  Gulin}.} \bibinfo{year}{2018}\natexlab{}.
\newblock \showarticletitle{{CatBoost}: Unbiased Boosting with Categorical
  Features}. In \bibinfo{booktitle}{\emph{Advances in Neural Information
  Processing Systems 31 (NeurIPS)}}. \bibinfo{publisher}{Curran Associates,
  Inc.}, \bibinfo{pages}{6639--6649}.
\newblock


\bibitem[Somepalli et~al\mbox{.}(2021)]%
        {somepalli2021saint}
\bibfield{author}{\bibinfo{person}{Gowthami Somepalli}, \bibinfo{person}{Micah
  Goldblum}, \bibinfo{person}{Avi Schwarzschild}, \bibinfo{person}{C.~Bayan
  Bruss}, {and} \bibinfo{person}{Tom Goldstein}.}
  \bibinfo{year}{2021}\natexlab{}.
\newblock \showarticletitle{{SAINT}: Improved Neural Networks for Tabular Data
  via Row Attention and Contrastive Pre-Training}.
\newblock \bibinfo{journal}{\emph{arXiv preprint arXiv:2106.01342}}
  (\bibinfo{year}{2021}).
\newblock


\bibitem[Ucar et~al\mbox{.}(2021)]%
        {ucar2021subtab}
\bibfield{author}{\bibinfo{person}{Talip Ucar}, \bibinfo{person}{Ehsan
  Hajiramezanali}, {and} \bibinfo{person}{Lindsay Edwards}.}
  \bibinfo{year}{2021}\natexlab{}.
\newblock \showarticletitle{{SubTab}: Subsetting Features of Tabular Data for
  Self-Supervised Representation Learning}. In
  \bibinfo{booktitle}{\emph{Advances in Neural Information Processing Systems
  34 (NeurIPS)}}. \bibinfo{publisher}{Curran Associates, Inc.},
  \bibinfo{pages}{18853--18865}.
\newblock


\bibitem[van~den Oord et~al\mbox{.}(2018)]%
        {oord2018representation}
\bibfield{author}{\bibinfo{person}{A{\"a}ron van~den Oord},
  \bibinfo{person}{Yazhe Li}, {and} \bibinfo{person}{Oriol Vinyals}.}
  \bibinfo{year}{2018}\natexlab{}.
\newblock \showarticletitle{Representation Learning with Contrastive Predictive
  Coding}.
\newblock \bibinfo{journal}{\emph{arXiv preprint arXiv:1807.03748}}
  (\bibinfo{year}{2018}).
\newblock


\bibitem[Yoon et~al\mbox{.}(2020)]%
        {yoon2020vime}
\bibfield{author}{\bibinfo{person}{Jinsung Yoon}, \bibinfo{person}{Yao Zhang},
  \bibinfo{person}{James Jordon}, {and} \bibinfo{person}{Mihaela van~der
  Schaar}.} \bibinfo{year}{2020}\natexlab{}.
\newblock \showarticletitle{{VIME}: Extending the Success of Self- and
  Semi-Supervised Learning to Tabular Domain}. In
  \bibinfo{booktitle}{\emph{Advances in Neural Information Processing Systems
  33 (NeurIPS)}}. \bibinfo{publisher}{Curran Associates, Inc.}
\newblock


\bibitem[Zhang et~al\mbox{.}(2023)]%
        {zhang2023fata}
\bibfield{author}{\bibinfo{person}{Dongyu Zhang}, \bibinfo{person}{Liang Wang},
  \bibinfo{person}{Xin Dai}, \bibinfo{person}{Shubham Jain},
  \bibinfo{person}{Junpeng Wang}, \bibinfo{person}{Yujie Fan},
  \bibinfo{person}{Chin-Chia~Michael Yeh}, \bibinfo{person}{Yan Zheng},
  \bibinfo{person}{Zhongfang Zhuang}, {and} \bibinfo{person}{Wei Zhang}.}
  \bibinfo{year}{2023}\natexlab{}.
\newblock \showarticletitle{{FATA-Trans}: Field and Time-Aware Transformer for
  Sequential Tabular Data}. In \bibinfo{booktitle}{\emph{Proceedings of the
  32nd ACM International Conference on Information and Knowledge Management
  (CIKM)}}. \bibinfo{publisher}{ACM}, \bibinfo{pages}{3247--3256}.
\newblock
\href{https://doi.org/10.1145/3583780.3614879}{doi:\nolinkurl{10.1145/3583780.3614879}}


\bibitem[Zhu et~al\mbox{.}(2023)]%
        {zhu2023xtab}
\bibfield{author}{\bibinfo{person}{Bingzhao Zhu}, \bibinfo{person}{Xingjian
  Shi}, \bibinfo{person}{Nick Erickson}, \bibinfo{person}{Mu Li},
  \bibinfo{person}{George Karypis}, {and} \bibinfo{person}{Mahsa Shoaran}.}
  \bibinfo{year}{2023}\natexlab{}.
\newblock \showarticletitle{{XTab}: Cross-table Pretraining for Tabular
  Transformers}. In \bibinfo{booktitle}{\emph{Proceedings of the 40th
  International Conference on Machine Learning (ICML)}}.
\newblock


\end{thebibliography}

\end{document}